\documentclass[runningheads]{llncs}
\usepackage[T1]{fontenc}
\usepackage{graphicx}
\begin{document}
\title{Performance comparison of medical image classification systems using TensorFlow Keras, PyTorch, and JAX}

\author{
Merjem Bećirović \and
Amina Kurtović \and
Nordin Smajlović \and
Medina Kapo \and
Amila Akagić
}

\institute{
Faculty of Electrical Engineering, University of Sarajevo,\\
Computer Science and Informatics,\\
71000 Sarajevo, Bosnia and Herzegovina\\
\email{\{mbecirovic3, akurtovic4, nsmajlovic1, mkapo2, aakagic\}@etf.unsa.ba}
}

\maketitle
\begin{abstract}
Medical imaging plays a vital role in early disease diagnosis and monitoring. Specifically, blood microscopy offers valuable insights into blood cell morphology and the detection of hematological disorders. In recent years, deep learning-based automated classification systems have demonstrated high potential in enhancing the accuracy and efficiency of blood image analysis. However, a detailed performance analysis of specific deep learning frameworks appears to be lacking. This paper compares the performance of three popular deep learning frameworks, TensorFlow with Keras, PyTorch, and JAX, in classifying blood cell images from the publicly available BloodMNIST dataset. The study primarily focuses on inference time differences, but also classification performance for different image sizes. The results reveal variations in performance across frameworks, influenced by factors such as image resolution and framework-specific optimizations. Classification accuracy for JAX and PyTorch was comparable to current benchmarks, showcasing the efficiency of these frameworks for medical image classification. 
\end{abstract}
\section{Introduction}
\label{sec:introduction}
Medical imaging is crucial in modern healthcare, enabling early diagnosis and monitoring of various diseases. Blood microscopy has gained significant attention for its ability to provide valuable insights into blood cell morphology, characteristics, and the diagnosis of hematological disorders and infections. Automated classification systems powered by deep learning have shown remarkable promise in improving accuracy and efficiency in analyzing blood microscopy images, making them an essential tool for pathologists and clinicians \cite{bloodmnist_2}.

Despite the widespread adoption of deep learning in medical imaging, selecting the most efficient framework for deployment remains a challenge due to differences in execution speed, optimization capabilities, and support for hardware accelerators. This paper addresses the problem of performance variability by comparing TensorFlow with Keras, PyTorch, and JAX, three leading deep learning frameworks, under identical model and training configurations. By evaluating inference speed and classification accuracy on blood cell microscopy images, we aim to identify which framework offers the most efficient execution for medical image classification tasks. Understanding these trade-offs is essential for researchers and practitioners seeking to deploy robust and scalable systems in clinical environments where both accuracy and speed are critical.

Deep learning has shown remarkable progress over the past decade, driving the creation of numerous frameworks. Among these, TensorFlow, PyTorch, and the more recent JAX have established themselves as leading tools in both industry and academia.

\subsection{TensorFlow and Keras}
Released by Google Brain in 2015, TensorFlow is a leading framework known for its scalability and performance optimizations. It represents computations as directed graphs, where operations are nodes and tensors flow along edges. TensorFlow efficiently maps computations to devices (CPU, GPU, TPU) and handles memory allocation and kernel execution \cite{tensorflow2015-whitepaper}. TensorFlow supports both static and dynamic graph execution, and it can be integrated with XLA\cite{openxla}, machine learning compiler that optimizes linear algebra, providing improvements in execution speed and memory usage. 
TensorFlow supports distributed training, custom kernels, and efficient data pipelines for parallelized data loading \cite{tensorflow2015-whitepaper}. 

Keras, TensorFlow's high-level API, simplifies model building, training, and evaluation \cite{chollet2015keras}. From Keras 3.0, it also supports PyTorch and JAX, but this paper uses TensorFlow’s Keras API for its stability and tight integration with TensorFlow's performance optimization features.

\subsection{PyTorch}
Introduced by Meta AI in 2016, PyTorch was designed to balance usability and speed, offering an imperative, Pythonic programming style that simplifies model development and debugging while ensuring efficiency and support for hardware accelerators like GPUs \cite{pytorch_paper}. Before PyTorch, frameworks like TensorFlow used static computational graphs for performance optimization, sacrificing ease of use and flexibility. Dynamic eager execution frameworks like Chainer \cite{Tokui2015Chainer} and Torch \cite{Collobert2002Torch} improved usability but often compromised performance or used faster but less expressive programming languages \cite{pytorch_paper}. PyTorch addressed this gap by combining dynamic eager execution, automatic differentiation, and GPU acceleration, maintaining performance comparable to the fastest static graph frameworks. PyTorch’s careful optimization of execution, memory management, and parallel processing further enhanced its efficiency, setting a new standard for deep learning usability. Its success influenced TensorFlow to adopt eager execution in version 2.x to improve flexibility and user experience. Over time, PyTorch also addressed production needs, bridging the gap between research and deployment with tools like TorchScript and TorchServe. In PyTorch, JIT and XLA are available for optimization but need to be explicitly enabled, with JIT focusing on performance improvements and XLA primarily used for TPU acceleration.

\subsection{JAX}
JAX, a 2018 research project developed by Google, is a Python library focused on accelerator-driven array computation and program transformation, aimed at enabling high-performance numerical computing and large-scale machine learning \cite{jax2018github}. By leveraging the XLA compiler, JAX optimizes code for various hardware architectures, allowing for flexible composition of computational kernels. JAX’s name, 'just after execution', reflects its approach of compiling functions after tracing their initial execution in Python \cite{jax_paper}.

A core feature of JAX is its JIT compilation, which transforms Python functions into efficient, accelerable code. JAX achieves this by tracing the function’s execution and recording all operations performed on the inputs. These operations are then reduced to a sequence of primitive functions that can be efficiently compiled and executed \cite{jax_jit_2025}. 

JAX’s transformations are designed to work with functionally pure Python functions, meaning all inputs are passed through parameters and all outputs are returned as results. This ensures that the functions can be efficiently compiled and optimized, as there are no unpredictable side-effects. More details on JAX’s purity constraints are discussed in \cite{jax_jit_2025}. 

\section{Related work}
\label{sec:related_work}
Few studies have compared TensorFlow, PyTorch, and JAX, with JAX being relatively new. One study \cite{pinns_comparing} evaluated Physics-Informed Neural Networks (PINNs) across these frameworks, showing JAX’s 23.68x speed-up over TensorFlow V2 and PyTorch for simpler problems, although TensorFlow outperformed JAX in large-scale tasks with higher batch sizes and more parameters. In comparisons between TensorFlow and PyTorch, a study \cite{s22228872} found that PyTorch excelled in training and execution speeds (25.5\% and 77.7\%, respectively), while TensorFlow offered higher accuracy and flexibility, making it better suited for precision-focused tasks. A comparison between TensorFlow V1 and PyTorch for single-GPU training \cite{Dai2021} highlighted the importance of GPU kernel execution time in training speed, with TensorFlow’s V1 static graph optimizations having minimal impact compared to PyTorch’s dynamic graph, suggesting that users should consider factors beyond graph type when choosing frameworks. A study \cite{SimmonsChance2019ACO} found similar accuracy for both frameworks, but TensorFlow had higher training time and lower memory usage, while PyTorch was better for prototyping and TensorFlow suited tasks requiring custom features. Finally, a study \cite{article_769457} found that TensorFlow performed best with small images, while PyTorch excelled with large images due to better memory management.

Many studies on microscopy image classification use BloodMNIST, with most employing variations of transformers to achieve state-of-the-art results. The study \cite{bloodmnist_1} evaluates quantum convolutional neural networks (QNNs) for multi-class classification on the BloodMNIST dataset, comparing them with traditional models. Research presented in \cite{bloodmnist_2} achieves a new benchmark accuracy of 97.90\% on BloodMNIST using the Vision Transformer model. A hybrid model combining CNNs with Random Forest and XGBoost is proposed in \cite{bloodmnist_3}, demonstrating superior performance and faster inference times. A novel generative model for semi-supervised disease classification is introduced in \cite{bloodmnist_4}, surpassing KL divergence-based approaches on five benchmarks, including BloodMNIST. Study \cite{bloodmnist_5} demonstrates the Medical Vision Transformer (MedViT) outperforms classical methods on MedMNIST. Compact Convolutional Transformers (CCTs) achieve 92.49\% accuracy on a small blood cell dataset in \cite{bloodmnist_6}, showing promise in addressing data scarcity. Lastly, \cite{bloodmnist_7} presents EfficientSwin, a hybrid model combining EfficientNet and Swin Transformer, achieving 98.14\% accuracy using 224x224 image resolutions.

\section{Methodology}
\label{sec:methodology}

\subsection{Dataset}
\begin{figure}[htbp]
\centering
\includegraphics[width=0.8\textwidth]{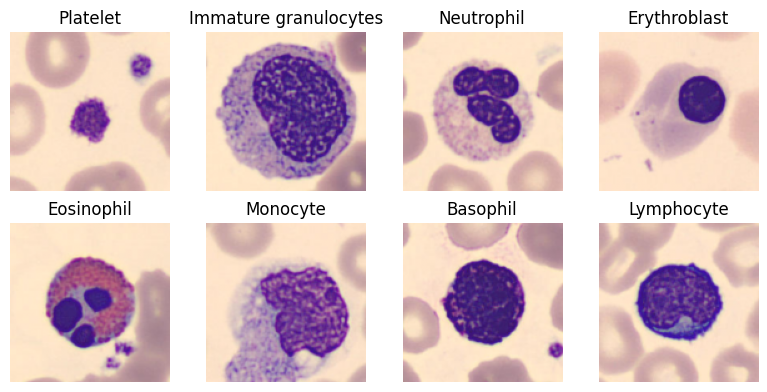}  
\caption{Example images from the BloodMNIST dataset with corresponding class labels.}
\label{fig:example_images}
\end{figure}

The dataset used for the experiments is BloodMNIST\cite{bloodmnist}, part of the MedMNIST\cite{medmnistv1,medmnistv2} collection of medical image datasets. BloodMNIST contains 17,092 2D blood cell microscopy images from healthy individuals. It is publicly available in .npz format at resolutions of 28×28, 64×64, 128×128, and 224×224 on the official MedMNIST page. The dataset is pre-split into training, validation, and testing subsets in a 7:1:2 ratio. This dataset is designed for a classification task with 8 distinct classes, each representing a type of blood cell, as summarized in Table \ref{tab:label_distribution}. The original pixel values, ranging from 0 to 255, were normalized to the range 0 to 1 for the experiments. Due to computational constraints, this paper focuses on performance analysis using images resized to 28×28 and 64×64 resolutions. Example images with their corresponding labels are shown in Figure \ref{fig:example_images}.

\begin{table}[ht]
\centering
\caption{Class distribution in the BloodMNIST train dataset}
\begin{tabular}{|c|l|c|c|}
\hline
\textbf{Label} & \textbf{Class name} & \textbf{N (total 11959)} & \textbf{Percentage (\%)} \\
\hline
0 & Basophil & 852 & 7.12 \\
\hline
1 & Eosinophil & 2181 & 18.24 \\
\hline
2 & Erythroblast & 1085 & 9.07 \\
\hline
3 & Immature granulocytes \footnote{myelocytes, metamyelocytes, promyelocytes} & 2026 & 16.94 \\
\hline
4 & Lymphocyte & 849 & 7.10 \\
\hline
5 & Monocyte & 993 & 8.30 \\
\hline
6 & Neutrophil & 2330 & 19.48 \\
\hline
7 & Platelet & 1643 & 13.74 \\
\hline
\end{tabular}
\label{tab:label_distribution}
\end{table}

\subsection{Architecture and training}
Our approach utilizes a custom ResNet\cite{DBLP:journals/corr/HeZRS15} inspired convolutional neural network (CNN) designed to balance computational efficiency and performance. The model begins with an initial convolutional layer with 32 filters to extract low-level features, followed by Batch Normalization and ReLU activation to stabilize training and introduce non-linearity. The core of the architecture comprises six residual blocks, each consisting of two convolutional layers with 3×3 kernels and skip connections. These residual connections mitigate the vanishing gradient problem, allowing deeper networks to train effectively.

\begin{figure}[htbp]
\centering
\includegraphics[width=1\textwidth]{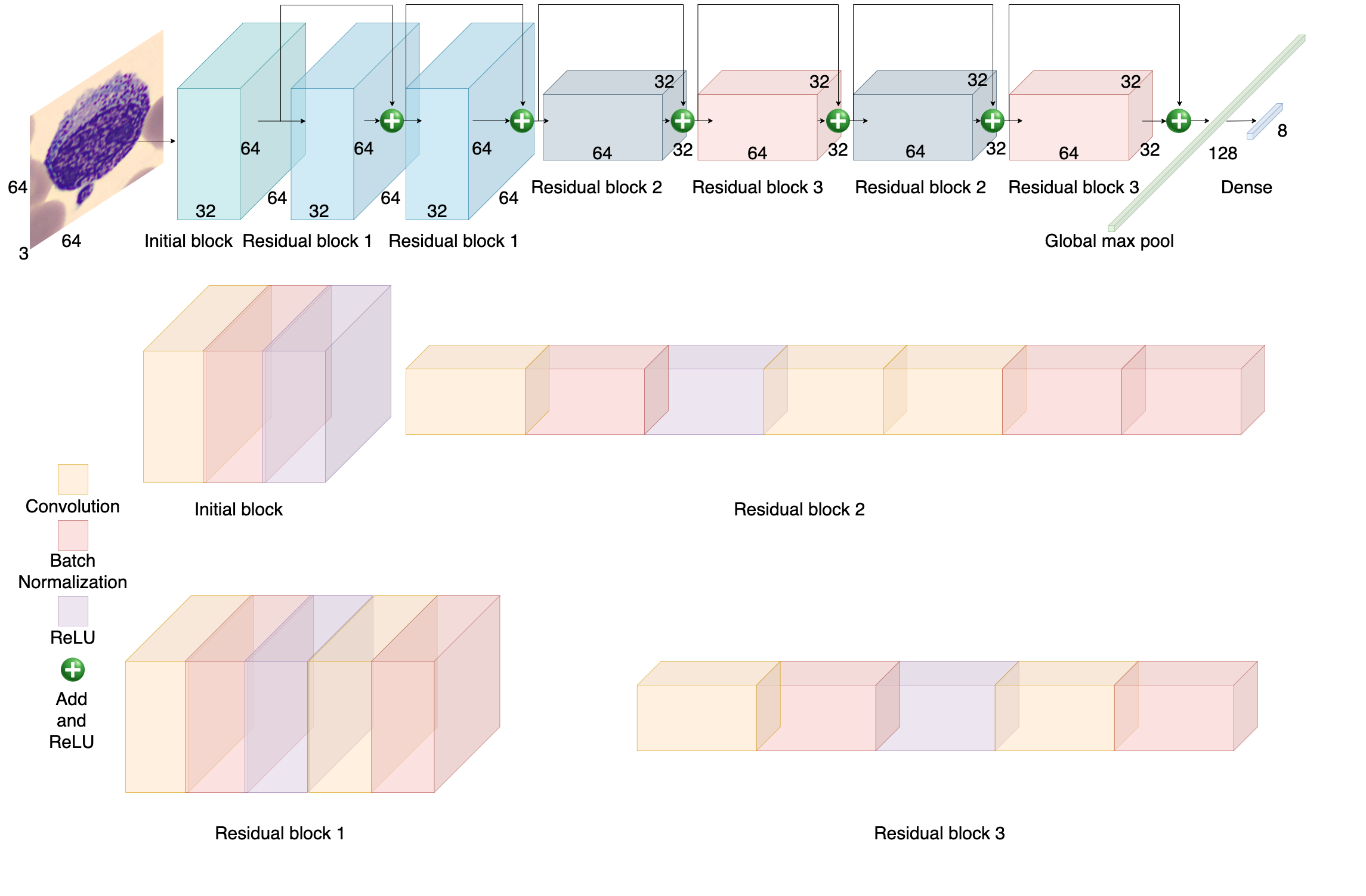}  
\caption{Custom ResNet architecture used for image classification.}
\label{fig:arch}
\end{figure}

The number of filters in the residual blocks doubles after every two blocks (from 32 to 64 to 128) to progressively learn higher-level features, while spatial dimensions are reduced using a stride of 2. A global max-pooling layer reduces the feature map into a vector representation, which is passed to a fully connected layer with a softmax activation to produce class probabilities for the eight blood cell types. VarianceScaling initialization is employed for weights to improve convergence, and Batch Normalization is applied after each convolution to normalize activations. The architecture is tailored for efficient training and robust feature extraction while maintaining model simplicity.  The architecture is illustrated in Figure \ref{fig:arch}.

For training, we compile the model with the Adam optimizer, categorical cross-entropy loss, and accuracy as the evaluation metric. The model is trained for 20 epochs using a batch size of 128, with a validation set to monitor the model's performance during training. This training configuration allows the model to converge efficiently while avoiding overfitting and ensures a balance between model complexity and computational cost.

\subsection{Inference time evaluation}
To evaluate and compare the inference time across Keras, JAX, and PyTorch, the prediction time was measured over 10 iterations on the 3,421 test images for each framework. The time for each iteration was recorded, and the average prediction time was computed for each model. This allowed the assessment of the relative efficiency of the models in generating predictions across the different frameworks.

For our experiments, we utilized the Tesla T4 GPU, which is optimized for deep learning model inference and features 40 streaming multiprocessors and 16GB of GDDR6 memory.

\section{Results}
\label{sec:results}
\subsection{Inference times}

The results in Table \ref{tab:inference_time_comparison} show distinct differences in inference times between the frameworks for both image sizes. For the 28x28 images, PyTorch demonstrates the fastest average inference time, significantly outperforming TensorFlow with Keras and JAX. This highlights PyTorch's superior efficiency for smaller image sizes. For the 64x64 images, JAX provides the fastest inference time, outperforming PyTorch, while Keras shows the slowest performance. The trend shows that as the image size increases, the difference in inference times between frameworks becomes slightly less pronounced, with JAX and PyTorch showing more comparable performance for larger image sizes. 

Despite JAX’s reputation for being highly optimized and efficient, it did not outperform PyTorch for the 28x28 image classification task. This can be attributed to several factors. Firstly, JAX utilizes JIT compilation to optimize operations, which can yield faster performance in certain tasks. However, for smaller image sizes like 28x28, the overhead introduced by JIT compilation might outweigh its benefits, particularly when the operations involved are relatively simple. On the other hand, PyTorch, which is highly optimized for a broad range of operations, may be more efficient for smaller images due to its built-in optimization strategies. Additionally, the specific model architecture and operations used could have influenced the performance, with PyTorch possibly better handling these smaller images. Furthermore, the smaller image size might have resulted in less effective parallelization, a factor that PyTorch, known for its strong parallel computation support, may have managed more efficiently than JAX. Therefore, while JAX is a powerful framework, it may not always be the fastest choice for all image sizes or tasks, particularly when dealing with smaller images where other frameworks like PyTorch can demonstrate superior performance.

\begin{table}[ht]
\centering
\caption{Inference Time Comparison Across Frameworks and Image Dimensions}
\label{tab:inference_time_comparison}
\begin{tabular}{|c|c|c|}
\hline
\textbf{Framework} & \textbf{Average time [28x28] (s)} & \textbf{Average time [64x64] (s)} \\ \hline
TensorFlow Keras              & 0.8985  & 2.0182       \\ \hline
PyTorch            & \textbf{0.3032}  & 1.5961     \\ \hline
JAX                & 0.3620  & \textbf{1.2692}    \\ \hline
\end{tabular}
\end{table}

\subsection{Classification performance}

Although the models across all three frameworks were designed with the same architecture and settings, the results shown in Tables \ref{tab:metrics_comparison_28x28} and \ref{tab:metrics_comparison_64x64} reveal interesting variations in their performance. For the 28x28 images, JAX demonstrated a consistent edge over both TensorFlow with Keras and PyTorch across all evaluation metrics, including accuracy, macro, and weighted averages for precision, recall, and F1-score. PyTorch slightly outperformed TensorFlow with Keras in certain areas, highlighting subtle differences between the frameworks despite using identical architectures and training setups.

For the 64x64 images, the results remain competitive. PyTorch achieved the highest accuracy (98.22\%), closely followed by JAX (97.9\%) and TensorFlow with Keras (97.63\%). PyTorch also led in macro and weighted averages for precision, recall, and F1-score, while JAX exhibited strong performance in precision. TensorFlow with Keras, although slightly behind, still maintained robust results across all metrics. These differences may arise from framework-specific implementations, such as optimization strategies or computational precision, which can influence model training dynamics. Overall, the results suggest that as image size increases, the gap in performance slightly narrows, yet minor variations between frameworks persist, underscoring the nuanced impact of framework-specific characteristics on model performance.
\begin{table}[ht]
\centering
\caption{Detailed Metrics Comparison for TensorFlow with Keras, PyTorch, and JAX (28x28 Image Size)}
\label{tab:metrics_comparison_28x28}
\begin{tabular}{|c|c|c|c|}
\hline
\textbf{Metric}              & \textbf{TensorFlow Keras} & \textbf{PyTorch} & \textbf{JAX}   \\ \hline
Accuracy                     & 0.9442         & 0.9486           & \textbf{0.9570} \\ \hline
Macro Avg Precision          & 0.9383         & 0.9463           & \textbf{0.9545} \\ \hline
Macro Avg Recall             & 0.9368         & 0.9417           & \textbf{0.9489} \\ \hline
Macro Avg F1-Score           & 0.9375         & 0.9438           & \textbf{0.9514} \\ \hline
Weighted Avg Precision       & 0.9444         & 0.9492           & \textbf{0.9576} \\ \hline
Weighted Avg Recall          & 0.9442         & 0.9486           & \textbf{0.9570} \\ \hline
Weighted Avg F1-Score        & 0.9442         & 0.9487           & \textbf{0.9571} \\ \hline
\end{tabular}
\end{table}

\begin{table}[ht]
\centering
\caption{Detailed Metrics Comparison for TensorFlow with Keras, PyTorch, and JAX (64x64 Image Size)}
\label{tab:metrics_comparison_64x64}
\begin{tabular}{|c|c|c|c|}
\hline
\textbf{Metric}              & \textbf{TensorFlow Keras} & \textbf{PyTorch} & \textbf{JAX}   \\ \hline
Accuracy                     & 0.9763         & \textbf{0.9822}  & 0.9790         \\ \hline
Macro Avg Precision          & 0.9754         & \textbf{0.9829}  & 0.9803         \\ \hline
Macro Avg Recall             & 0.9758         & \textbf{0.9820}  & 0.9771         \\ \hline
Macro Avg F1-Score           & 0.9755         & \textbf{0.9824}  & 0.9785         \\ \hline
Weighted Avg Precision       & 0.9764         & \textbf{0.9822}  & 0.9792         \\ \hline
Weighted Avg Recall          & 0.9763         & \textbf{0.9822}  & 0.9790         \\ \hline
Weighted Avg F1-Score        & 0.9763         & \textbf{0.9821}  & 0.9790         \\ \hline
\end{tabular}
\end{table}

\section{Conclusion}
\label{sec:conclusion}

This study focused on comparing TensorFlow with Keras, PyTorch, and JAX in terms of inference time and classification performance for 28x28 and 64x64 blood microscopy images. In terms of inference time, PyTorch was the fastest for 28x28 images, while JAX outperformed the other frameworks for 64x64 images, demonstrating its strength in handling larger images efficiently. It is worth noting that JAX's JIT compilation, which optimizes computation, may introduce some overhead during the initial setup, affecting performance for smaller image sizes. Despite using the same architecture and settings across all frameworks, JAX showed superior performance on classification metrics for 28x28 images, while PyTorch outperformed the other two for 64x64 images. These results suggest that while inference time plays a critical role in framework selection, the differences in classification performance may arise from factors such as weight initialization, optimization algorithms, and JIT compilation overhead. One of the primary limitations of this study lies in the resolution of the analyzed images (28×28 and 64×64 pixels), which are significantly lower than those typically encountered in clinical digital microscopy. While this choice enables faster experimentation and highlights inference differences across frameworks, it limits the direct applicability of the findings to real-world medical imaging scenarios. Nevertheless, this work can be viewed as a feasibility study, offering insights that are likely to generalize to larger resolutions, particularly in terms of framework behavior and scaling trends. One way to work with higher resolution images would be to use super-resolution methods. Namely, by using super-resolution methods such as Enhanced Deep Super-Resolution (EDSR), Efficient Sub-Pixel Convolutional Neural Network (ESPCN), Super-Resolution Convolutional Neural Network (SRCNN) or Super-Resolution Generative Adversarial Network (SRGAN), small-sized and low-resolution images could be brought to better quality and dimensions, which would make this study more complete and general.

\bibliographystyle{splncs04}
\bibliography{literature}

\begin{thebibliography}{10}
\providecommand{\url}[1]{\texttt{#1}}
\providecommand{\urlprefix}{URL }
\providecommand{\doi}[1]{https://doi.org/#1}

\bibitem{tensorflow2015-whitepaper}
Abadi, M., Agarwal, A., Barham, P., Brevdo, E., Chen, Z., Citro, C., Corrado, G.S., Davis, A., Dean, J., Devin, M., Ghemawat, S., Goodfellow, I., Harp, A., Irving, G., Isard, M., Jia, Y., Jozefowicz, R., Kaiser, L., Kudlur, M., Levenberg, J., Man\'{e}, D., Monga, R., Moore, S., Murray, D., Olah, C., Schuster, M., Shlens, J., Steiner, B., Sutskever, I., Talwar, K., Tucker, P., Vanhoucke, V., Vasudevan, V., Vi\'{e}gas, F., Vinyals, O., Warden, P., Wattenberg, M., Wicke, M., Yu, Y., Zheng, X.: {TensorFlow}: Large-scale machine learning on heterogeneous systems (2015), \url{https://www.tensorflow.org/}, software available from tensorflow.org

\bibitem{bloodmnist}
Acevedo, A., Merino, A., Alférez, S., Ángel Molina, Boldú, L., Rodellar, J.: A dataset of microscopic peripheral blood cell images for development of automatic recognition systems. Data in Brief  \textbf{30},  105474 (2020)

\bibitem{jax_jit_2025}
Authors, J.: Jax official documentation (2024), \url{https://jax.readthedocs.io/en/latest/jit-compilation.html}, accessed: 2025-01-19

\bibitem{bloodmnist_3}
Aydın, M., Kuş, Z., Akçelik, Z.K.: A hybrid cnn-tree based model for enhanced image classification performance. In: 2024 32nd Signal Processing and Communications Applications Conference (SIU). pp.~1--4 (2024). \doi{10.1109/SIU61531.2024.10600973}

\bibitem{pinns_comparing}
Bafghi, R.A., Raissi, M.: Comparing {PINN}s across frameworks: {JAX}, tensorflow, and pytorch. In: ICLR 2024 Workshop on AI4DifferentialEquations In Science (2024), \url{https://openreview.net/forum?id=BPFzolSSrI}

\bibitem{jax2018github}
Bradbury, J., Frostig, R., Hawkins, P., Johnson, M.J., Leary, C., Maclaurin, D., Necula, G., Paszke, A., Vander{P}las, J., Wanderman-{M}ilne, S., Zhang, Q.: {JAX}: composable transformations of {P}ython+{N}um{P}y programs (2018), \url{http://github.com/jax-ml/jax}

\bibitem{chollet2015keras}
Chollet, F., et~al.: Keras. \url{https://keras.io} (2015)

\bibitem{Collobert2002Torch}
Collobert, R., Bengio, S., Mariéthoz, J.: Torch: A modular machine learning software library. Technical Report Idiap-RR-02-46, Idiap (2002), https://www.idiap.ch/en/scientific-reports/2002/techreports-idiap-rr-02-46

\bibitem{Dai2021}
Dai, H., Peng, X., Shi, X., He, L., Xiong, Q., Jin, H.: Reveal training performance mystery between tensorflow and pytorch in the single gpu environment. Science China Information Sciences  \textbf{65} (2021). \doi{10.1007/s11432-020-3182-1}, \url{https://doi.org/10.1007/s11432-020-3182-1}

\bibitem{jax_paper}
Frostig, R., Johnson, M., Leary, C.: Compiling machine learning programs via high-level tracing (2018), \url{https://mlsys.org/Conferences/doc/2018/146.pdf}

\bibitem{bloodmnist_6}
Gao, A.K.: More for less: Compact convolutional transformers enable robust medical image classification with limited data. arXiv preprint arXiv:2307.00213  (2023)

\bibitem{bloodmnist_2}
Halder, A., Gharami, S., Sadhu, P., Singh, P.K., Wo{\'z}niak, M., Ijaz, M.F.: Implementing vision transformer for classifying 2d biomedical images. Scientific Reports  \textbf{14}(1),  12567 (2024)

\bibitem{DBLP:journals/corr/HeZRS15}
He, K., Zhang, X., Ren, S., Sun, J.: Deep residual learning for image recognition. CoRR  \textbf{abs/1512.03385} (2015), \url{http://arxiv.org/abs/1512.03385}

\bibitem{bloodmnist_1}
Khmelnytskyi, A., Stirenko, S., Gordienko, Y.: Hybrid neural networks for medical image classification. In: Manoharan, S., Tugui, A., Baig, Z. (eds.) Proceedings of 4th International Conference on Artificial Intelligence and Smart Energy. pp. 462--474. Springer Nature Switzerland, Cham (2024)

\bibitem{bloodmnist_5}
Liu, Y.: Medical image classification based on transformer model and ordinal loss

\bibitem{s22228872}
Novac, O.C., Chirodea, M.C., Novac, C.M., Bizon, N., Oproescu, M., Stan, O.P., Gordan, C.E.: Analysis of the application efficiency of tensorflow and pytorch in convolutional neural network. Sensors  \textbf{22}(22) (2022). \doi{10.3390/s22228872}, \url{https://www.mdpi.com/1424-8220/22/22/8872}

\bibitem{openxla}
OpenXLA: Xla (accelerated linear algebra) documentation. \url{https://openxla.org/xla} (2021), accessed: 2023-01-19

\bibitem{pytorch_paper}
Paszke, A., Gross, S., Massa, F., Lerer, A., Bradbury, J., Chanan, G., Killeen, T., Lin, Z., Gimelshein, N., Antiga, L., Desmaison, A., K{\"{o}}pf, A., Yang, E.Z., DeVito, Z., Raison, M., Tejani, A., Chilamkurthy, S., Steiner, B., Fang, L., Bai, J., Chintala, S.: Pytorch: An imperative style, high-performance deep learning library. CoRR  \textbf{abs/1912.01703} (2019), \url{http://arxiv.org/abs/1912.01703}

\bibitem{bloodmnist_7}
Patel, T., El-Sayed, H., Sarker, M.K.: Efficientswin: A hybrid model for blood cell classification with saliency maps visualization. In: 2024 35th Conference of Open Innovations Association (FRUCT). pp. 544--551. IEEE (2024)

\bibitem{SimmonsChance2019ACO}
SimmonsChance, HollidayMark, A.: A comparison of two popular machine learning frameworks. Journal of Computing Sciences in Colleges  (2019), \url{https://api.semanticscholar.org/CorpusID:231142130}

\bibitem{Tokui2015Chainer}
Tokui, S., Oono, K., Hido, S., Clayton, J.: Chainer: A next-generation open source framework for deep learning. In: Proceedings of Workshop on Machine Learning Systems (LearningSys) in The Twenty-ninth Annual Conference on Neural Information Processing Systems (NIPS). Montreal, Canada (2015), https://arxiv.org/abs/1511.05852

\bibitem{medmnistv1}
Yang, J., Shi, R., Ni, B.: Medmnist classification decathlon: A lightweight automl benchmark for medical image analysis. In: IEEE 18th International Symposium on Biomedical Imaging (ISBI). pp. 191--195 (2021)

\bibitem{medmnistv2}
Yang, J., Shi, R., Wei, D., Liu, Z., Zhao, L., Ke, B., Pfister, H., Ni, B.: Medmnist v2-a large-scale lightweight benchmark for 2d and 3d biomedical image classification. Scientific Data  \textbf{10}(1), ~41 (2023)

\bibitem{article_769457}
Yapıcı, M.M., Topaloğlu, N.: Performance comparison of deep learning frameworks. Computers and Informatics  \textbf{1}(1),  1--11 (2021)

\bibitem{bloodmnist_4}
Zhang, Y., Li, C., Liu, Z., Li, M.: Semi-supervised disease classification based on limited medical image data. IEEE Journal of Biomedical and Health Informatics  (2024)

\end{thebibliography}

\end{document}